\documentclass[letterpaper, 10 pt, conference]{ieeeconf}  

\IEEEoverridecommandlockouts                              

\title{\LARGE \bf
Design and Control of SQUEEZE: A Spring-augmented QUadrotor for intEractions with the Environment to squeeZE-and-fly}

\author{Karishma Patnaik, Shatadal Mishra, Seyed Mostafa Rezayat Sorkhabadi and Wenlong Zhang$^{*}$
\thanks{The authors are with The Polytechnic School, Ira A. Fulton Schools of Engineering, Arizona State University, Mesa, AZ, 85212, USA. Email: {\tt\small $\{$kpatnaik, smishr13, wenlong.zhang$\}$@asu.edu}.}%
\thanks{$^{*}$Address all correspondence to this author. The first two authors have equal contribution to this work.}
\thanks{This work was supported in part by the Salt River Project.}%
}

\usepackage{textcomp}
\usepackage{overpic}
\usepackage{gensymb}
\usepackage{url}
\usepackage{amsmath}
\usepackage{graphicx}
\usepackage{amssymb}
\usepackage{mathtools}
\usepackage{xcolor}
\usepackage{float}
\usepackage{cite}
\usepackage{multirow}
\usepackage[caption=false]{subfig}
\usepackage{caption}
\usepackage{booktabs}
\usepackage{siunitx}
\usepackage{stfloats}
\usepackage{MnSymbol}
\newcommand{\revised}[1]{\textcolor{black}{#1}}
\newcommand{\revisedb}[1]{\textcolor{black}{#1}}

\makeatletter
\let\NAT@parse\undefined
\makeatother
\usepackage{hyperref}

\DeclarePairedDelimiter\norm{\lVert}{\rVert}

\begin{document}

\graphicspath{{figures/}}
\maketitle
 \thispagestyle{empty}
\pagestyle{empty}

\begin{abstract}
This paper presents the design and control of a novel
quadrotor with a variable geometry to physically interact
with cluttered environments and fly through narrow gaps and passageways. This compliant quadrotor with passive morphing capabilities is designed using torsional springs at every arm hinge to allow for rotation \revised{driven by} external forces. We derive the dynamic model of this variable geometry quadrotor (SQUEEZE), and develop \revised{an} adaptive controller for trajectory tracking. The corresponding Lyapunov stability proof of attitude tracking is also presented. Further, an admittance controller is designed to account for change\revised{s} in yaw due to physical interactions with the environment. Finally, the proposed design is validated in flight tests \revised{with} two setups: a small gap and a passageway. The experimental results demonstrate the unique capability of the SQUEEZE in navigating through constrained narrow spaces.
\end{abstract}

\section{Introduction}\label{sec:1}

Recent work on quadrotor systems demonstrate their applications for \revised{challenging} tasks such as inspections of cluttered and occluded environments
\cite{P+12, V+19}, aerial grasping \cite{M+18} and contact-based navigation, where a quadrotor interacts physically with the environment while navigating through it \cite{B+13, airobots, R+07,B+12}. Flying through such cluttered environments often requires the quadrotor to traverse through gaps smaller than its size, while simultaneously ensuring successful missions.

For a rigid quadrotor, flying through constrained spaces leads to research topics such as executing aggressive maneuvers \cite{F+17} or performing real-time trajectory re-planning to avoid cluttered areas using computer vision techniques \cite{S+16}. 
Alternatively, a quadrotor can have an adaptive morphology \cite{MF16}, where a \revised{folding mechanism helps the quadrotor} fly through narrow apertures. In this context, Riviere et al. \cite{R+18} proposed a variable geometry quadrotor using an actuated elastic mechanism to traverse vertical gaps. 
Falanga et al. \cite{F+18} employed external actuators to fold the quadrotor arms around the main body. 
Another folding mechanism using an origami laminate structure was proposed \cite{LM19}. In these designs, additional actuators are needed, which increases the total weight of the system and decreases its power-to-weight ratio. 
In \cite{BM19}, Bucki and Mueller proposed a design for quadrotor morphing and demonstrated flights through holes using passive rotary joints for arms. 
Despite different quadrotor designs with adaptive morphologies, they \revised{choose} to
avoid interactions with the edges of passages \revised{and} holes while flying through them. These interactions bring significant challenges for vehicle stability when the quadrotor experiences unknown interactions from the edges.

\begin{figure}
    \centering
    \includegraphics[width = 0.35\textwidth]{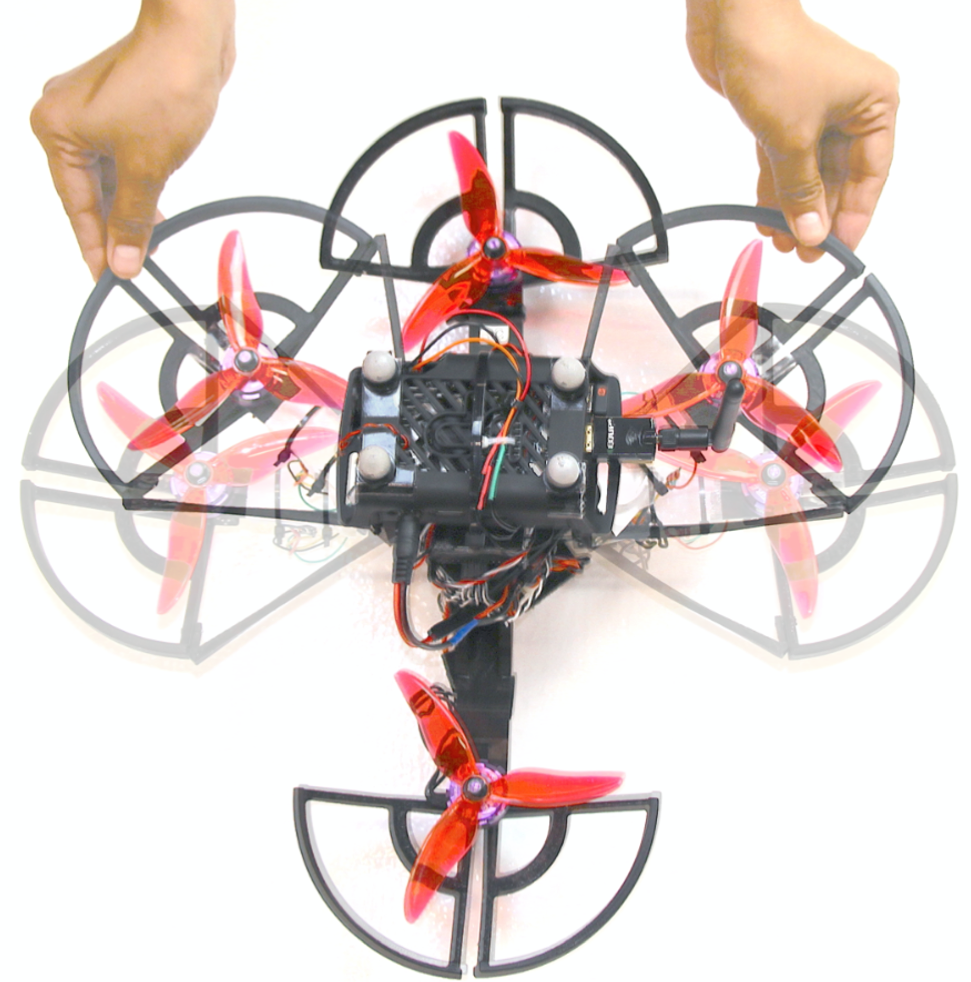}
    \caption{The quadrotor design proposed in this paper with a passive adaptive morphology when interacting with the environment. \revised{videolink: \url{https://youtu.be/0GEBscTuoDA}}} 
    \label{fig:first_fig}
    \vspace{-0.3in}
\end{figure}

Taking into consideration the aforementioned issues, we propose a new a new passive adaptive morphology design: a compliant quadrotor that can interact with environment and modify its shape in response to the external forces. We utilize torsional springs attached between the arms and the body frame, which enable in-plane rotary motion for these joints. Figure \ref{fig:first_fig} shows the vehicle in spring loaded and unloaded configurations. 
When in contact with an obstruction, the arms rotate about the hinge; this mechanical compliance \revised{results} in passive morphing and is critical for vehicle stability during physical interactions with the environment.

Compared to prior work on morphing quadrotors, the proposed design employs springs with negligible weight ($ < 2 \%$ of the total weight) for achieving adaptive morphology. Hence, the power-to-weight ratio is not significantly affected due to the added components. 
To the best of the authors’
knowledge, this is the first work in literature that showcases a compliant frame which adapts to narrow apertures by physically interacting with it. 

The rest of the paper is organized as follows: in Section \ref{sec:design}, we describe the details of the proposed design and its fabrication. In Section \ref{sec:modelling}, the modeling and system dynamics for a varying geometry are described. Section \ref{sec:control} discusses the proposed adaptive controller for trajectory tracking and the admittance controller for flight through passageways. Section \ref{sec:experiments} demonstrates experimental results to validate the proposed system for various testing  scenarios. Section \ref{sec:RNC} \revised{concludes the paper} and proposes some future work.

\begin{figure}
    \centering
    \includegraphics[width = 0.35\textwidth]{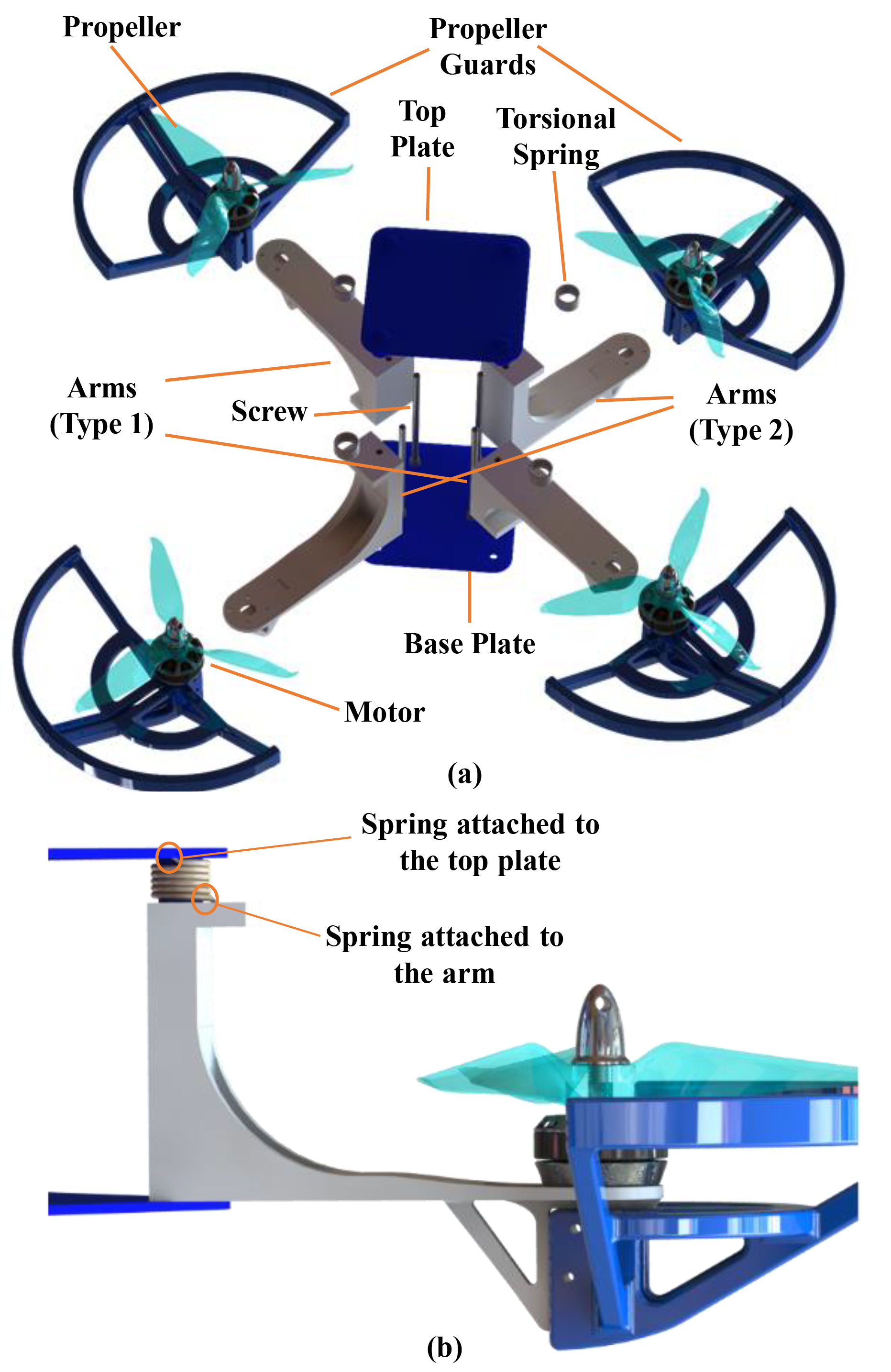}
    \caption{\revisedb{Overview of the SQUEEZE design.} (a) Exploded view, (b) Zoomed in view to show how arms are attached to the top plate using the torsional spring, which enables the arms to rotate relative to the top plate when forces are applied on them}
    \label{fig:breakOut}
\end{figure}

\section{Quadrotor Design}\label{sec:design}
The quadrotor frame and propeller guards are \revised {3D printed using polylactide}, making it lightweight and \revised {resilient} to impact forces experienced during interactions with the environment. The design consists of a base and top plate, four arms, four torsional springs and four propeller guards (Fig. \ref{fig:breakOut}a). The torsional springs are attached to the arm hinges, which connect the arms to the top plate. This will add a rotational degree of freedom (DoF) to each arm, giving it the ability to passively fold when in-plane external forces are exerted on the drone.
  This passive variable geometry design enables the quadrotor to fly through narrow gaps and passageways by physically interacting with the environment by rotating the arms. Propeller guards are attached to the outer side of the arms and are designed to ensure smooth transitions during such flights. The arms are designed in such a way that the adjacent motors and propellers are at different heights, thus preventing physical interference between the arms when folding occurs (arm types 1 and 2 in Fig. \ref{fig:breakOut}b). The selected spring is lightweight (4$g$), \revised {capable of} enough deflection (120\degree), and is \revised {compliant enough} (0.21 $N\cdot m/rad$) so that the arms can be folded without requiring the quadrotor to provide large thrust force when moving forward through narrow spaces. \revised{The current platform has a total width of 41 $cm$ and a folded width of 28 $cm$.}


\begin{table}[!b]
\centering
\caption{Nomenclature}
\label{table:nomen}
\resizebox{0.5\textwidth}{!}{%
    \begin{tabular}{p{1.5cm}p{5cm}}
        $m \in \mathbb{R}$ & point mass representing each pair of motor, propeller, and corresponding arm model \\
        $M \in \mathbb{R}$ & mass of the center piece \\
        $m_{t} \in \mathbb{R}$ & total mass of the entire system \\
        $x \in \mathbb{R}^3 $ & position of the vehicle in the inertial frame \\
        $v \in \mathbb{R}^3 $ & velocity of the vehicle in the inertial frame \\
        $J \in \mathbb{R}^{3 \times 3}$ & moment of inertia matrix in the body fixed frame \\
        $\Omega \in \mathbb{R}^3 $ & angular velocity in the body frame \\
        $R \in \mathsf{SO(3)}$ & rotation matrix from body frame to the inertial frame \\
        $cg \in \mathbb{R}^3$ & position of center of gravity of vehicle in the $\{q_1,q_2,q_3\}$ frame \\
        $\beta_1, \beta_2 \in \mathsf{S}^1$ & angles made by arms 2 and 4 respectively with the negative $q_1$ axis \\
        $f \in \mathbb{R}$ & control input of total thrust \\
        $\tau \in \mathbb{R}^3$ & control moments for roll, pitch and yaw \\
        $\gamma \in \mathbb{R}^4 $ & angles made by each arm with $q_2$ axis \\
        $r \in \mathbb{R}^4 $ & distance from each $m$ to the CG \\
        $h \in \mathbb{R}$ & height of motor pairs 2 and 4 in $q_3$ \\
        $k_\tau \in \mathbb{R}$ & spring constant 
        \end{tabular}
    }
\end{table}

\section{Model and System Dynamics}\label{sec:modelling}
\begin{figure}
    \centering
    \includegraphics[width = 0.45\textwidth]{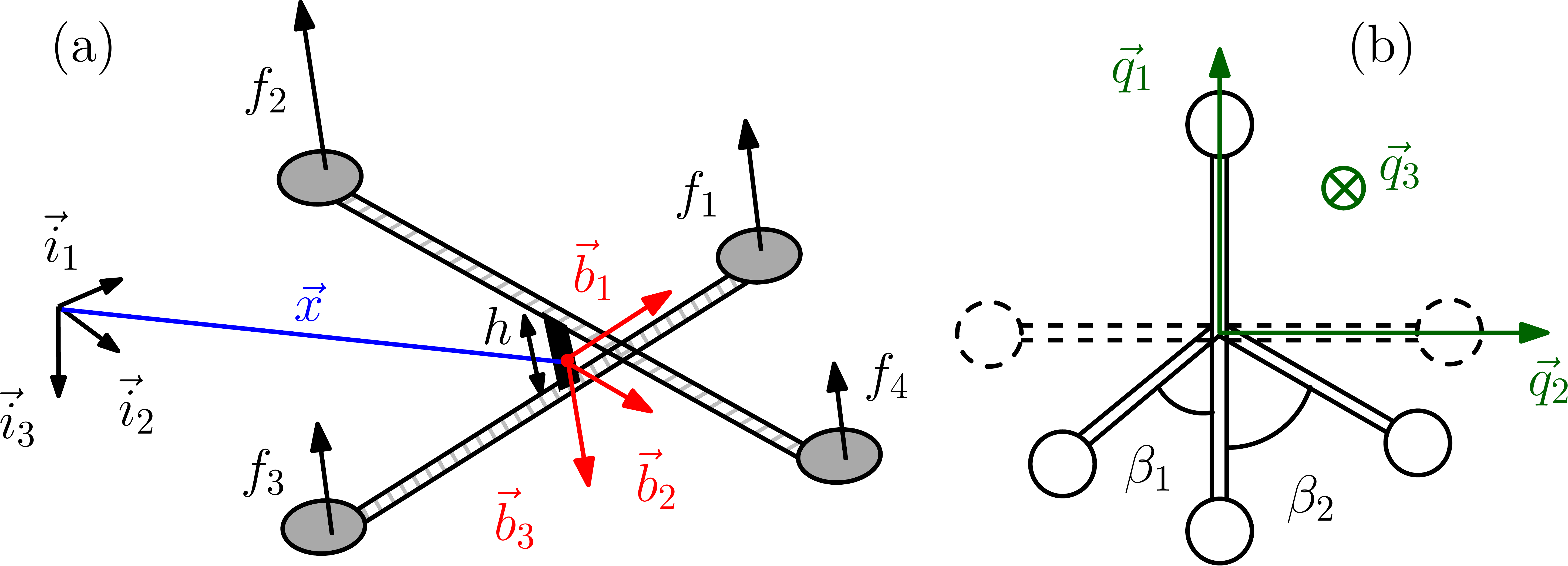}
    \caption{\revisedb{Coordinate systems for dynamic modeling.} In (a), $\{i_1,i_2,i_3\}$ denote the inertial frame and $\{b_1,b_2,b_3\}$ denote the body fixed frame respectively. Note that, arms $2$ and $4$ are placed a height lower than the pair $1$ and $3$. The thrust produced by each motor is perpendicular to the body $b_1-b_2$ plane and is denoted by $f_i$. $x$ denotes the position of the quadrotor's centre of mass at any instant of time in the inertial frame. In (b), the arms $2$ and $4$  bend $\beta_1$ and $\beta_2$ angles respectively depicting passive folding.}
    \label{fig:axis}
\end{figure}
In this section, we first describe the notations used in this paper and then specify the details of a simplified lumped mass model to calculate the position of center of gravity (CG) and the moment of inertia matrix at any time instant. 
For an asymmetric and varying CG, the mapping of control inputs to individual motor thrusts changes, so we also present the derivation of the control allocation matrix in this section. 
\subsection{Notation}\label{sec:notation}
For system dynamics and modeling, we define an inertial frame $\{i_1,i_2,i_3\}$ and a body frame $\{b_1,b_2,b_3 \}$. The origin of this body fixed frame is located at the CG of the vehicle as shown in Fig. \ref{fig:axis}. 
We also define a $\{q_1, q_2, q_3\}$ frame with its origin at the geometric centre of the top view of the frame in order to calculate the relative position of the varying CG with respect to it. The $q_1$ axis is parallel to the line connecting motors 1 and 3 when the corresponding arms are in their equilibrium positions. Similarly, the $q_2$ axis is defined perpendicular to $q_1$ \revised{and lies in $b_1-b_2$ plane}, as shown in Fig. \ref{fig:moi}. The $q_3$ axis is perpendicular to the $q_1 - q_2$ plane and is along the $b_3$ axis. Table \ref{table:nomen} lists the notations, and we use $(\cdot)_d$ to denote the desired quantity of $(\cdot)$ in this paper.

        

\begin{table*}[!t]
    \centering
    \caption{Moment of inertia calculation from angles $\beta_1, \beta_2$ (in $kg-m^2$)}
    \label{table:moi_calc}
    \begin{tabular}{ll}
    \hline 
      $J_{xx}$ &  $\frac{2}{5}MR^2 + M(cg_y^2+cg_z^2) + 2m(cg_y^2+cg_z^2) + m((-l\sin\beta_1 - cg_y)^2 +(h-cg_z)^2) + m((l\sin\beta_2 - cg_y)^2  +(h-cg_z)^2)$ \\[1ex]
      \hline
      $J_{yy}$ &  $ \frac{2}{5}MR^2 + M(cg_x^2+cg_z^2) + m((l-cg_x)^2 + cg_z^2) + m((-l-cg_x)^2 + cg_z^2) + m((-l\cos{\beta_1 - cg_x})^2 + (h-cg_z)^2)$ \\
      & $m((-l\cos{\beta_2 - cg_x})^2 + (h-cg_z)^2)$ \\[1ex]
      \hline
      $J_{zz}$ &  $\frac{2}{5}MR^2 + M(cg_x^2 + cg_y^2) + m(cg_y^2+(l-cg_x)^2) + m(cg_y^2+(-l-cg_x)^2) + m((-l\sin{\beta_1}-cg_y)^2+(-l\cos{\beta_1}-cg_x)^2)$ \\
         & $+ m((l\sin{\beta_2}-cg_y)^2+(-l\cos{\beta_2}-cg_x)^2)$ \\[1ex]
      \hline
      $J_{xy} = J_{yx}$ & $M cg_y cg_x -m cg_y(l-cg_x) - m cg_y(-l-cg_x) + m(-l\sin{\beta_1-cg_y})(-l\cos{\beta_1-cg_x})$ \\[1ex]
      \hline 
      $J_{yz} = J_{zy}$ & $M cg_y cg_z + 2m cg_y cg_z  + m(-l\sin{\beta_1 - cg_y})(h - cg_z) + m(l\sin{\beta_2} - cg_y)(h - cg_z)$ \\[1ex]
      \hline
      $J_{zx} = J_{xz}$ & $M cg_x cg_z - m(l-cg_x)cg_z - m(-l-cg_x)cg_z + m(-l\cos{\beta_1} - cg_x)(h - cg_z) + m(-l\cos{\beta_2} - cg_x)(h - cg_z)$ \\[1ex]
      \hline
    \end{tabular}
 \end{table*}
 
\subsection{Expression for Moment of Inertia}
\begin{figure}
    \centering
    \includegraphics[width=3.3in]{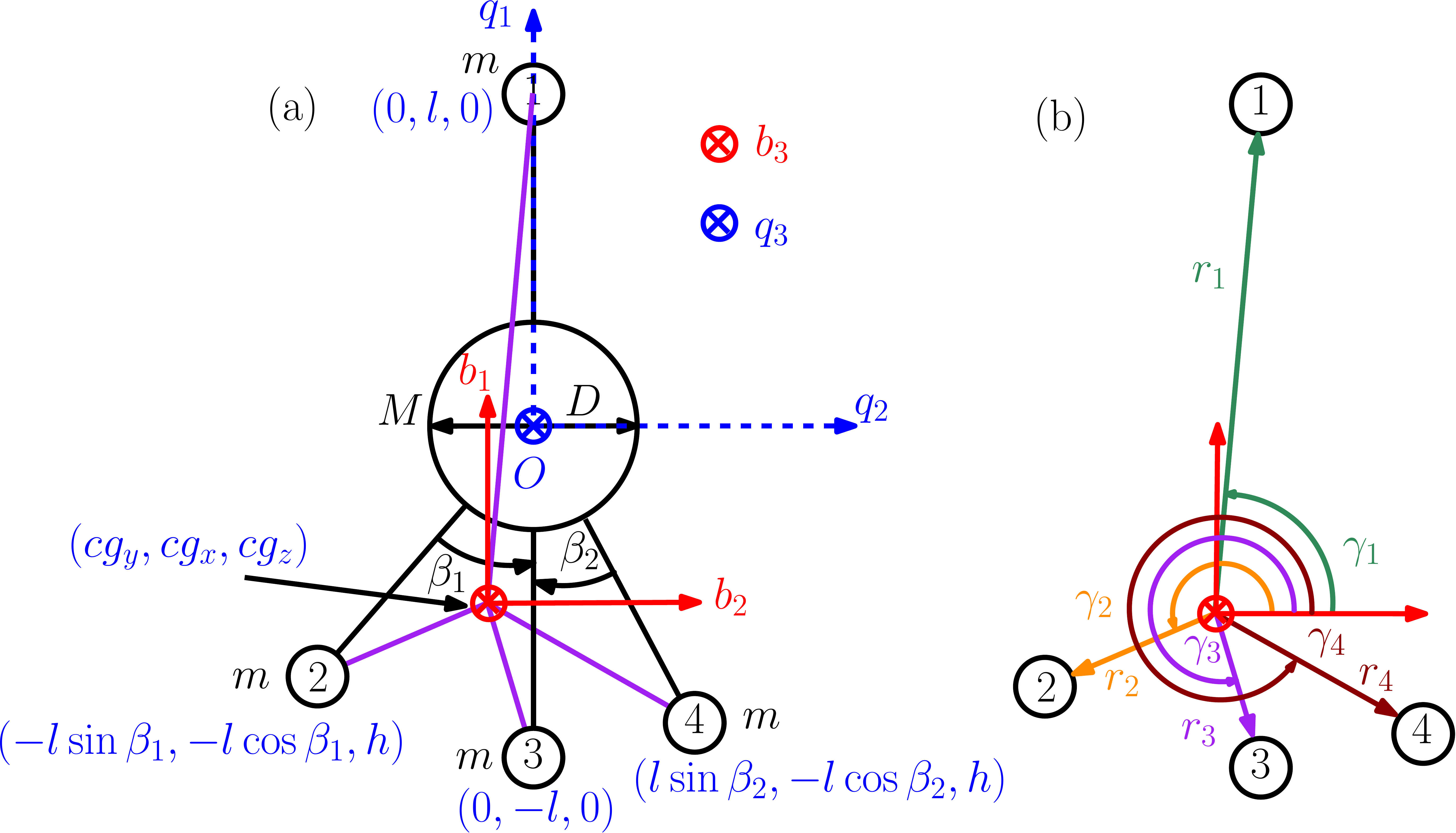}
    \caption{\revisedb{The moment of inertia model. The SQUEEZE is modeled as a} spherical dense center with mass $M$ and radius $R$, and point masses of mass $m$ located at a distance of $l$ from the center with arms 2 and 4 rotated $\beta_1$ and $\beta_2$ units about the center respectively. The motor pairs 2 and 4 are placed at a height $h$ along the $b_3$ axis. }
    \label{fig:moi}
    \vspace{-0.15in}
\end{figure}

Since the proposed quadrotor design has \revised{a varying} geometry while flying through passageways, the location of CG at any instant of time and the system's moment of inertia is necessary to compute the correct control moments for trajectory tracking. We model the system as an assemblage of a large sphere of mass $M$ with diameter $D$ and point masses of $m$ units at a distance of $l$ each from the center of the sphere. In addition, spheres $2$ and $4$ are at a height of $h$ units below the motor pairs of $1$ and $3$. We now calculate the inertia tensor of the whole body considering the $\{q_1,q_2,q_3 \}$ frame with origin $O$ as shown in Fig. \ref{fig:moi}.

Without loss of generality, let arms 2 and 4 make an angle of $\beta_1$ and $\beta_2$ with the negative $q_1$ axis respectively as shown in Fig.~\ref{fig:moi}. The coordinates for CG in $\{q_2,q_1,q_3\}$ denoted by ($cg_y,cg_x,cg_z$) as functions of $\beta_1, \beta_2$ are given by 
\begin{equation}\label{eqn:cg}
\centering
    \begin{aligned}
        cg = \big(-\mathcal{C}l(\sin \beta_1 - \sin \beta_2),  -\mathcal{C}l(\cos \beta_1 + \cos \beta_2), 2\mathcal{C}h \big),\\
    \end{aligned}
\end{equation}
where $\mathcal{C}=\big(\frac{m}{M+4m}\big)$. Let $J_{xx}, J_{yy}, J_{zz}$ denote the moment of inertia about the $b_1,b_2,b_3$ axis, respectively, then \revised{the time-varying moment of inertia of the system, $J(\beta_1(t),\beta_2(t))$}, can be written as (\ref{eqn:moi}) 
\begin{align}\label{eqn:moi}
     {J}(\beta_1(t),\beta_2(t)) = \begin{bmatrix} 
     J_{xx} & J_{xy} & J_{xz} \\ J_{yx} & J_{yy} & J_{yz} \\ J_{zx} &J_{zy} &J_{zz} \end{bmatrix},
 \end{align}
and the formulation of each term is described in Table \ref{table:moi_calc}.

The moment of inertia calculated using (\ref{eqn:moi}) is used to design an adaptive controller for trajectory tracking. This formulation is verified using a SolidWorks model as shown in Table \ref{table:moi} in the appendix. 
The parameter values of the lumped mass model for the current design are: $M ~=~710\textrm{g},m = 95\textrm{g}, D = 10\textrm{cm},l = 12.5\textrm{cm}, h = -3\textrm{cm}$.

\subsection{Defining Control Inputs and Control Allocation}
Assuming that the thrust produced by each propeller is directly controlled and that the direction of thrust generated is normal to the quadrotor plane, the total thrust is the sum of thrusts produced by each motor, that is, $f = \sum_{i=1}^4{f_i}$. Further, let $\tau_1, \tau_2$ and $\tau_3$ denote the pitch, roll and yaw moments respectively and $\gamma_i, i = 1,..,4$ denote the angles made by each arm with the positive $b_2$ axis as shown in Fig. \ref{fig:moi} (b). 
We can now calculate the moments about $b_1$ and $b_2$ axes as
 \begin{equation}\label{eqn:torque}
    \begin{aligned}
        \tau_1 ~=~& f_1 r_1 \sin{\gamma_1} + f_2 r_2 \sin{\gamma_2} \\
        & f_3 r_3 \sin{\gamma_3} + f_4 r_4 \sin{\gamma_4},\\
        \tau_2 ~=~& - f_1 r_1 \cos{\gamma_1} - f_2 r_2 \cos{\gamma_2} \\
        &- f_3 r_3 \cos{\gamma_3} - f_4 r_4 \cos{\gamma_4},\\
        \tau_3 ~=~& \sum_{i=1}^4 (-1)^i c_\tau f_i, 
    \end{aligned}
\end{equation}
where the torque generated by each propeller contributes to the total yaw moment and is equal to $(-1)^i c_\tau f_i$ for a fixed constant $c_\tau$.
We now define the control allocation matrix (CAM) of the total thrust $f$ and total moment $\tau$ as a mapping to individual motor thrusts, $f_1,f_2,f_3$ and $f_4$, as

\begin{equation}\label{eqn:cam}
    \begin{bmatrix}
        f \\ \tau_1 \\ \tau_2 \\ \tau_3
    \end{bmatrix}
    =
    \begin{bmatrix}
        1 & 1 & 1 & 1 \\ 
        r_1 s \gamma_1 & r_2 s \gamma_2 & r_3 s \gamma_3 & r_4 s \gamma_4 \\
        - r_1 c \gamma_1 & -r_2 c \gamma_2 & -r_3 c \gamma_3 & -r_4 c \gamma_4 \\
        -c_\tau & c_\tau & -c_\tau & c_\tau
    \end{bmatrix}
    \begin{bmatrix}
        f_1 \\ f_2 \\ f_3 \\ f_4
    \end{bmatrix},
\end{equation}
The determinant of this matrix is
\begin{equation*}
    \begin{aligned}
        det(\text{CAM}) ~=~ & 2c_\tau (r_1 r_2 \sin{(\gamma_1 - \gamma_2)} + r_1 r_4 \sin{(\gamma_1 - \gamma_4)} \\
        &+ r_2 r_3 \sin{(\gamma_3 - \gamma_2)} + r_3 r_4 \sin{(\gamma_4 - \gamma_3)}).
    \end{aligned}
\end{equation*}
We see that when arms $2,3,4$ are at the same location, 
this matrix is singular implying that we will lose some degrees of freedom for full attitude control and need to find solutions for a reduced order system. For this paper, we ensure that \revised{CAM} does not become singular at any instant of time. 

In the rest of the work, we assume that control inputs to the system are total thrust $f \in \mathbb{R}$ and torque $\tau \in \mathbb{R}^3$ and use (\ref{eqn:cam}) to calculate the individual thrust needed for each motor. 
\subsection{System Dynamics}
The dynamics of the quadrotor about the CG with control inputs $f$ and $\tau$ are given by
\begin{equation}\label{eqn:dynamics}
    \begin{aligned}
    m_t\dot{{v}} &= m_t g{e}_3 - f{R}{e}_3, \\
    \dot{{x}} &= {v}, \\ 
    \dot{{R}} &= {R}\hat{{\Omega}}, \\
    {J}\dot{{\Omega}} &= {\tau} - {\Omega} \times {J}{\Omega},
\end{aligned}
\end{equation}
where $g = 9.81 ~ ms^{-2}$ denotes the acceleration due to gravity, ${e}_3$ denotes the $i_3$ axis unit vector and the \textit{hat map} $\hat{\cdot}: \mathbb{R}^3 \xrightarrow{} \mathsf{SO(3)}$ is a symmetric matrix operator defined by the condition that $\hat{x}y = x \times y, ~\forall~ x,y \in \mathbb{R}^3$. \revised{Note that for the scope of this paper, friction dynamics is not considered.}

\begin{figure}
    \centering
    \includegraphics[width = 0.225\textwidth]{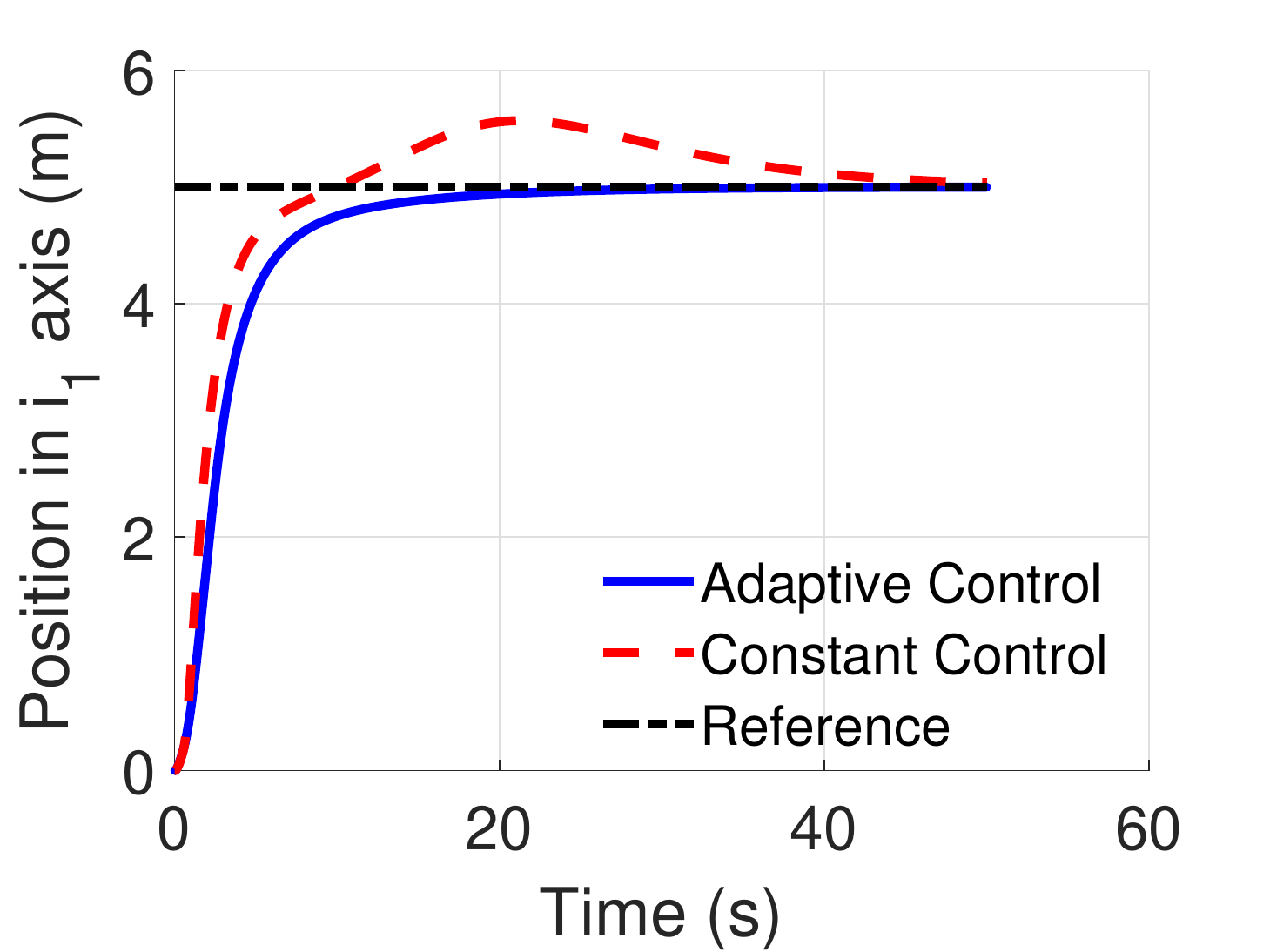}
    \includegraphics[width = 0.225\textwidth]{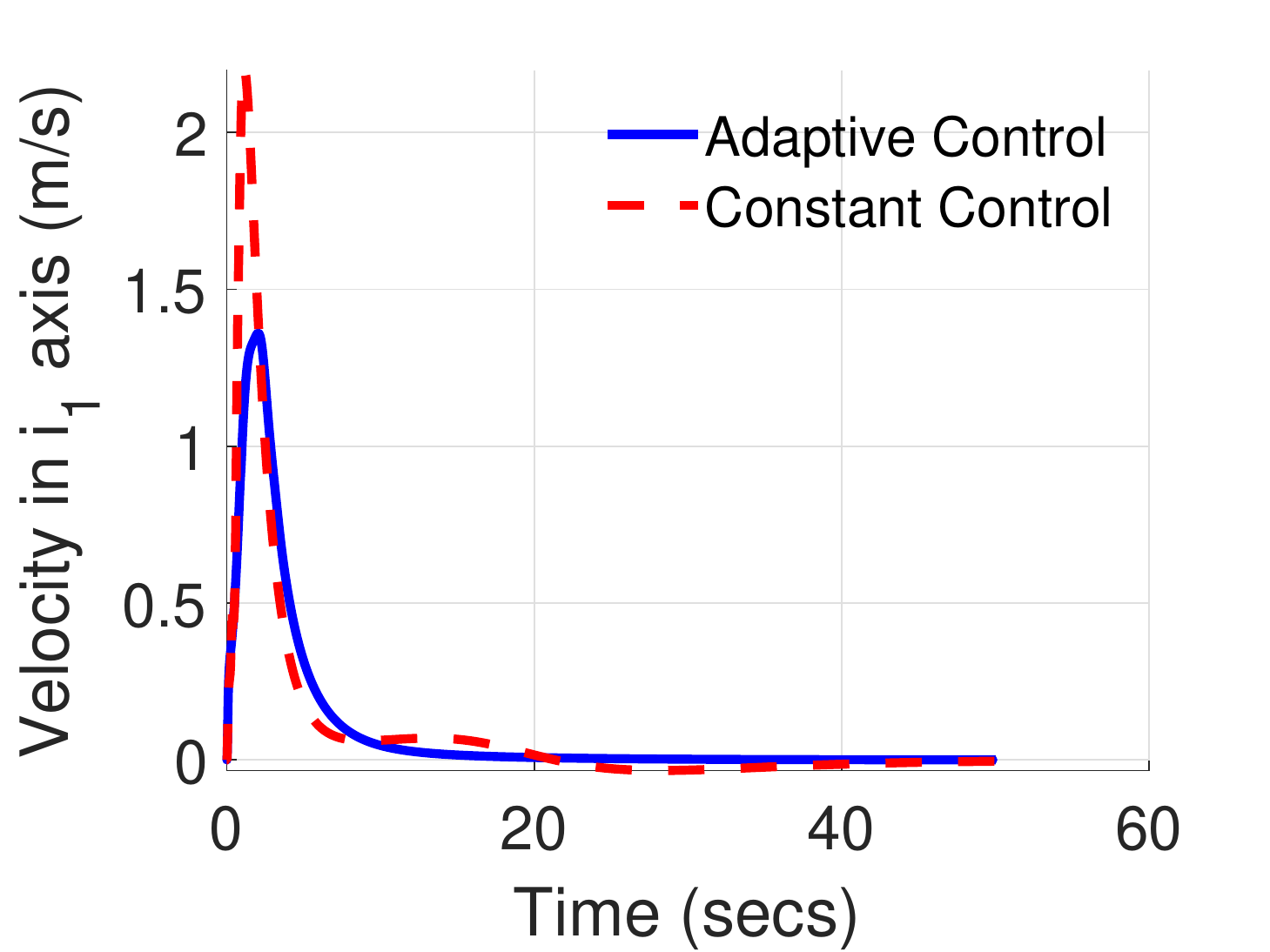}\\
    \includegraphics[width = 0.225\textwidth]{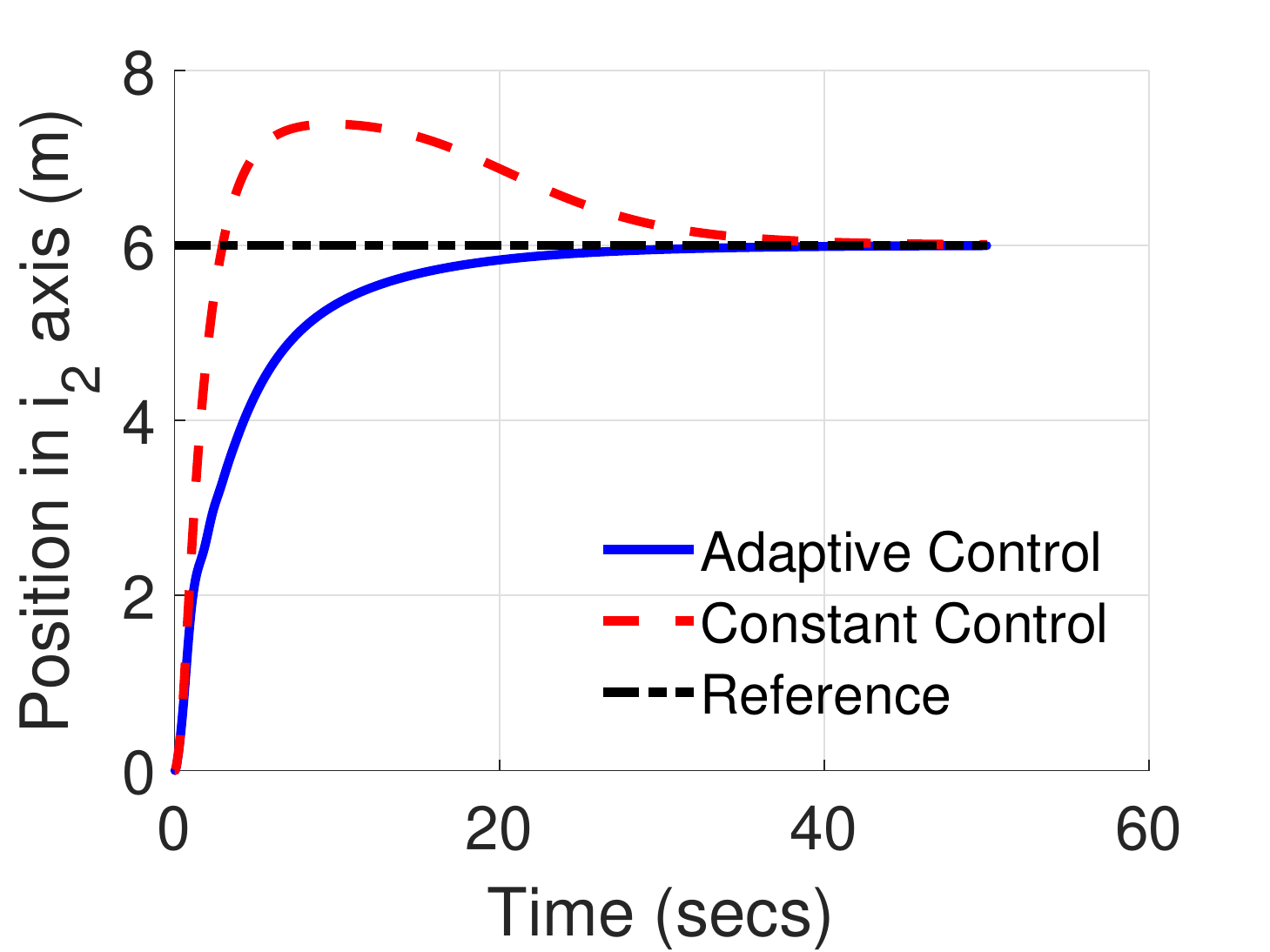}
    \includegraphics[width = 0.225\textwidth]{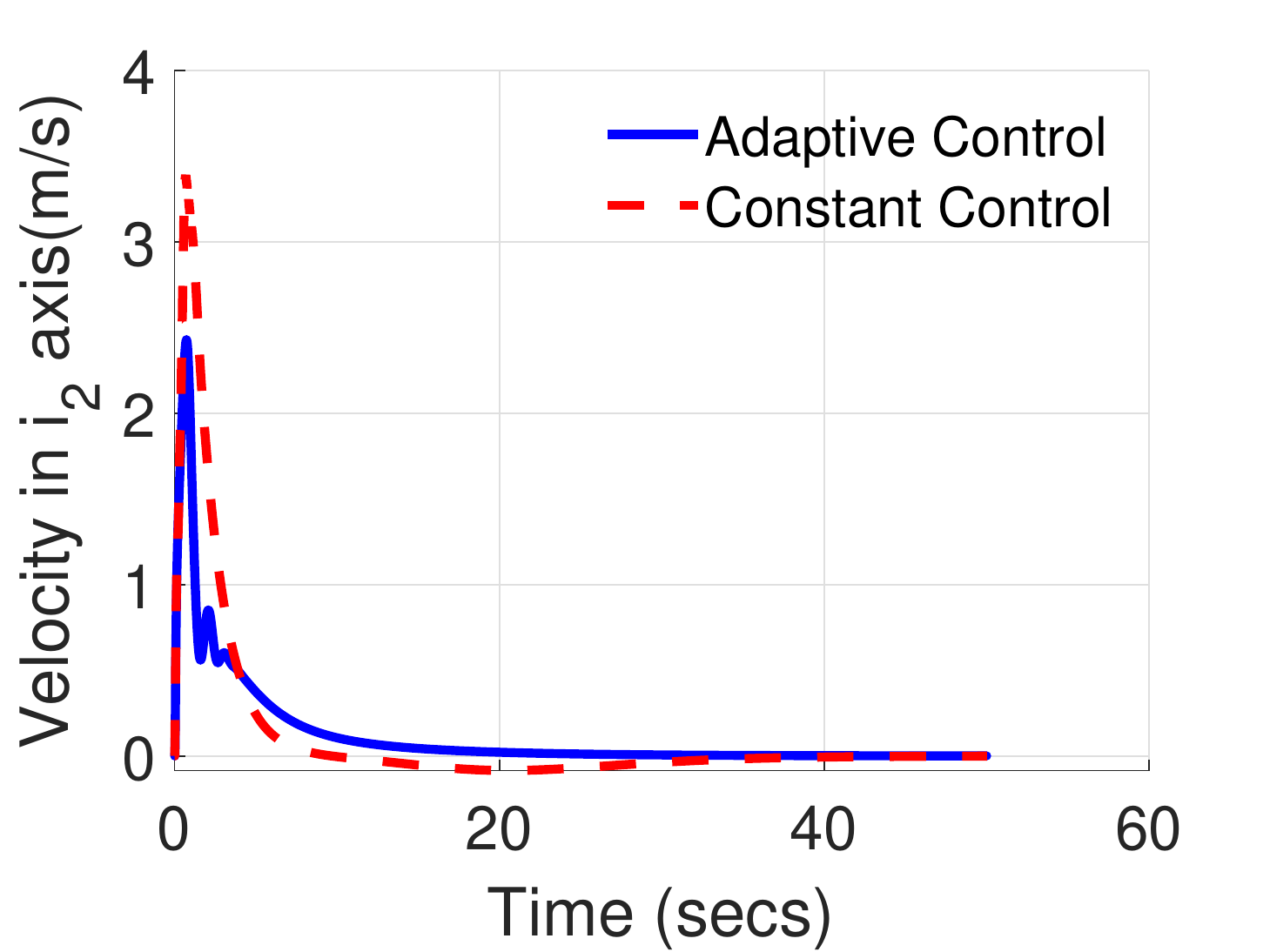}\\
    \includegraphics[width = 0.225\textwidth]{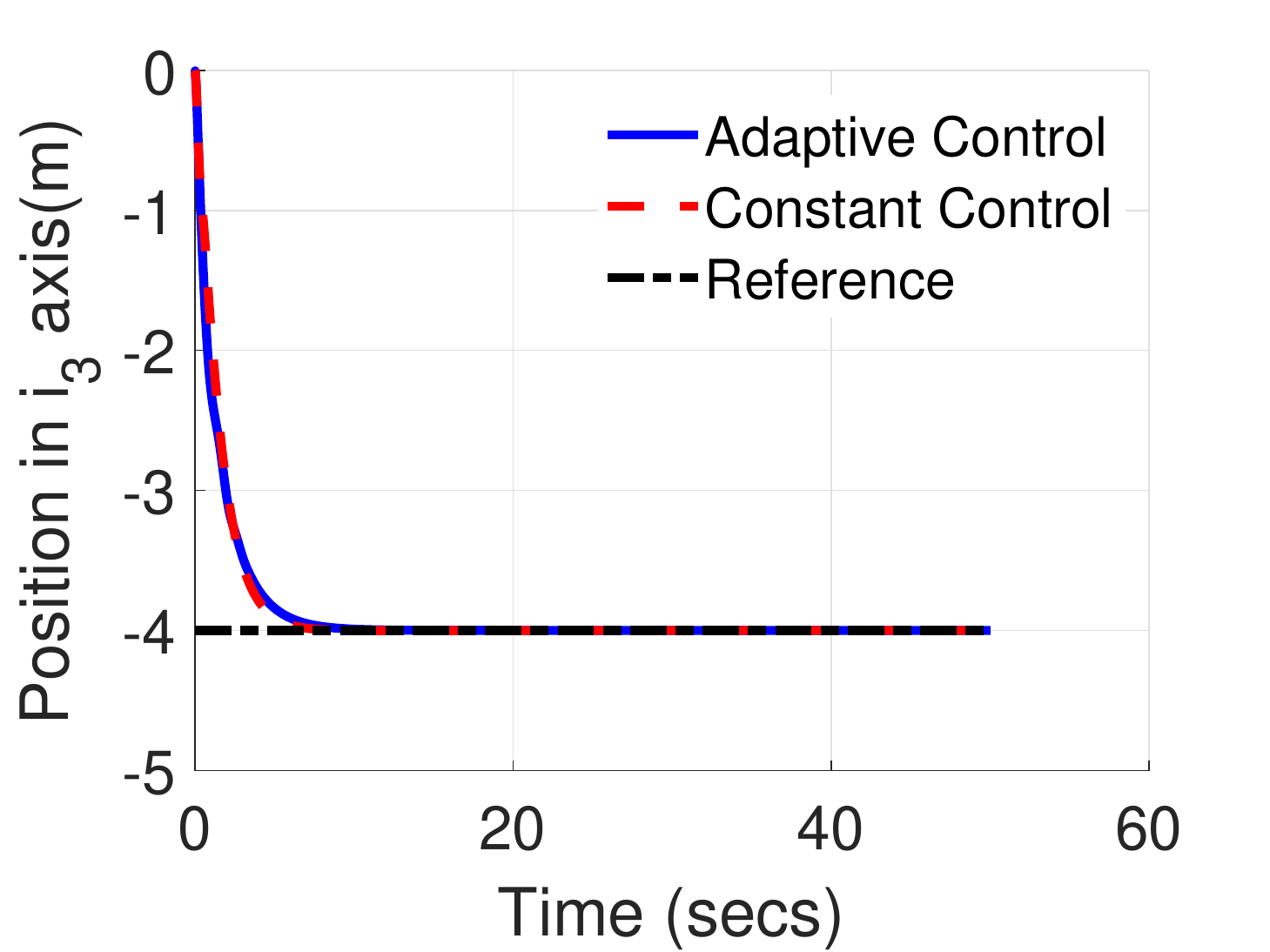}
    \includegraphics[width = 0.225\textwidth]{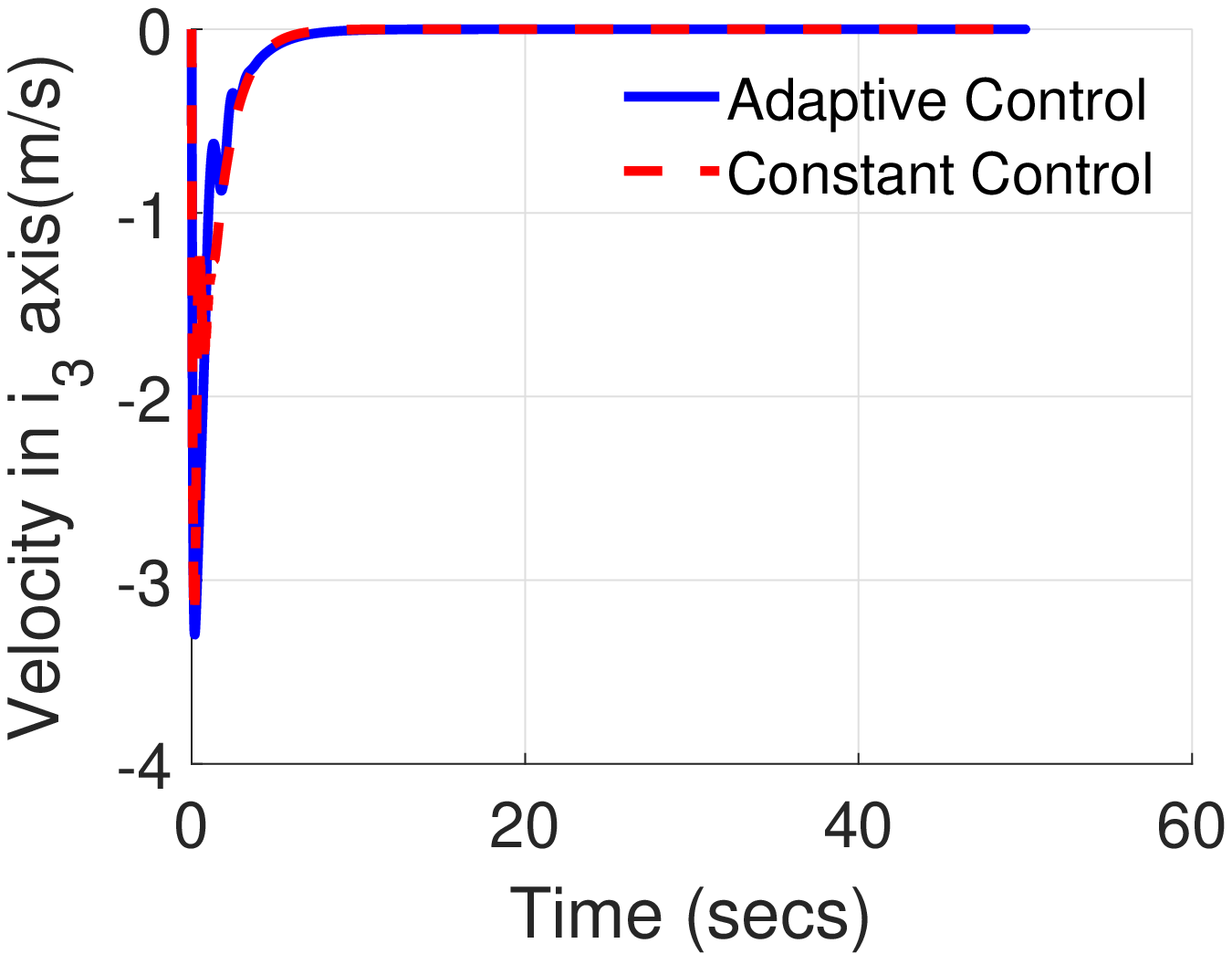}
    \caption{Simulation results showing trajectory tracking performance of a standard non-adaptive controller v.s. the proposed adaptive controller.} 
    \label{fig:sim_control}
\end{figure}

\section{Controller Design}\label{sec:control}
This section discusses the low-level attitude controller for trajectory tracking and how we leverage the design coupled with an admittance control for flying through passageways. 
\subsection{Adaptive Attitude Control}\label{sec:adaptivecontrol}
For trajectory tracking of a varying morphology, an adaptive attitude controller is proposed on the nonlinear configuration Lie group which accounts for the quadrotor's varying \revised{moment of inertia}. The desired $b_{1d},b_{2d}$ and $b_{3d}$ axes are chosen in a similar fashion as that of a standard quadrotor \cite{L+10}. Now, errors in $R$ and $\Omega$ are defined as \cite{L+10}
\begin{equation}
    \begin{aligned}
        e_\Omega ~=~ & \Omega - R^TR_d\Omega_d, \\
        e_R ~=~ & \frac{1}{2}(R^T_dR - R^TR_d)^\vee.
    \end{aligned}
\end{equation}
We choose the 
control moment $\tau \in \mathbb{R}^3$ as
\begin{equation}\label{eqn:control_tau}
    \begin{aligned}
        \tau = &~J(-k_R e_R - k_\Omega e_\Omega - k_{\Omega_d} \dot{e}_\Omega + \zeta_d) + \Omega \times J\Omega 
    \end{aligned}
\end{equation}
for any
positive constants $k_R, k_\Omega, k_{d\Omega}$ and $\zeta_d = -\hat{\Omega}R^TR_d\Omega_d + R^TR_d\dot{\Omega}_d$. We prove that with this controller, the tracking error of attitude dynamics will converge to zero asymptotically. \revised{The estimated angular acceleration, $\dot{\Omega}$, can be obtained by numerical differentiation of the estimated angular velocity. As, numerical differentiation results in noisy output, the estimated angular acceleration is not utilized for implementation purposes.}
\par
\textbf{Proof}: The asymptotic stability for the attitude error is given in the appendix. 

\subsubsection*{Simulations}
The comparison results of simulations for a case where $\beta_1 = 30^o$, $\beta_2 = 30^o$, $x_d = [5~5~-4]^T $ and $b_{1d} = [1 ~0 ~0]^T $ with the proposed adaptive controller and a standard controller \cite{L+10} \revised{which does not account for the varying ${J}(\beta_1(t),\beta_2(t))$} are shown in Fig. \ref{fig:sim_control}. It is noticed that for a short distance, the non-adaptive controller does not deviate significantly from the desired trajectory. However, if the morphing is persistent for a longer duration, with a non-adaptive control, there is significant deviation from the desired trajectory, whereas the adaptive controller achieves accurate tracking. This is attributed to the shifted CG and the varying moment of inertia which is not accounted by a non-adaptive controller to generate the appropriate $f$ and $\tau$ control signals.  

\begin{figure}
    \centering
    \includegraphics[width=0.45\textwidth]{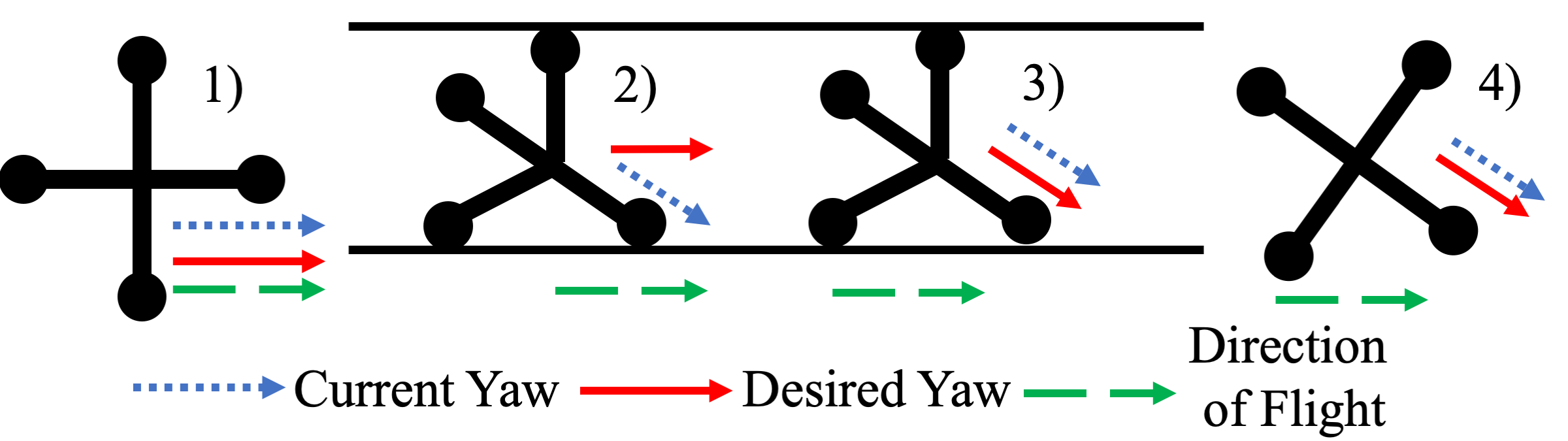} 
    \caption{Schematic showing how $\psi_d$ is updated to account for physical interactions: (1) The SQUEEZE enters the passageway, (2) the arms bend and current yaw changes due to the interaction forces,  (3) admittance controller updates the desired yaw so it converges to the current yaw, and (4) vehicle maintains this yaw throughout to the end of the passageway to comply with the interactions.}
    \label{fig:adm_schematic}
    \vspace{-0.1in}
\end{figure}

\subsection{Admittance Control}\label{admittancectrl}
\begin{figure}
    \centering
    \includegraphics[width = 3.3 in]{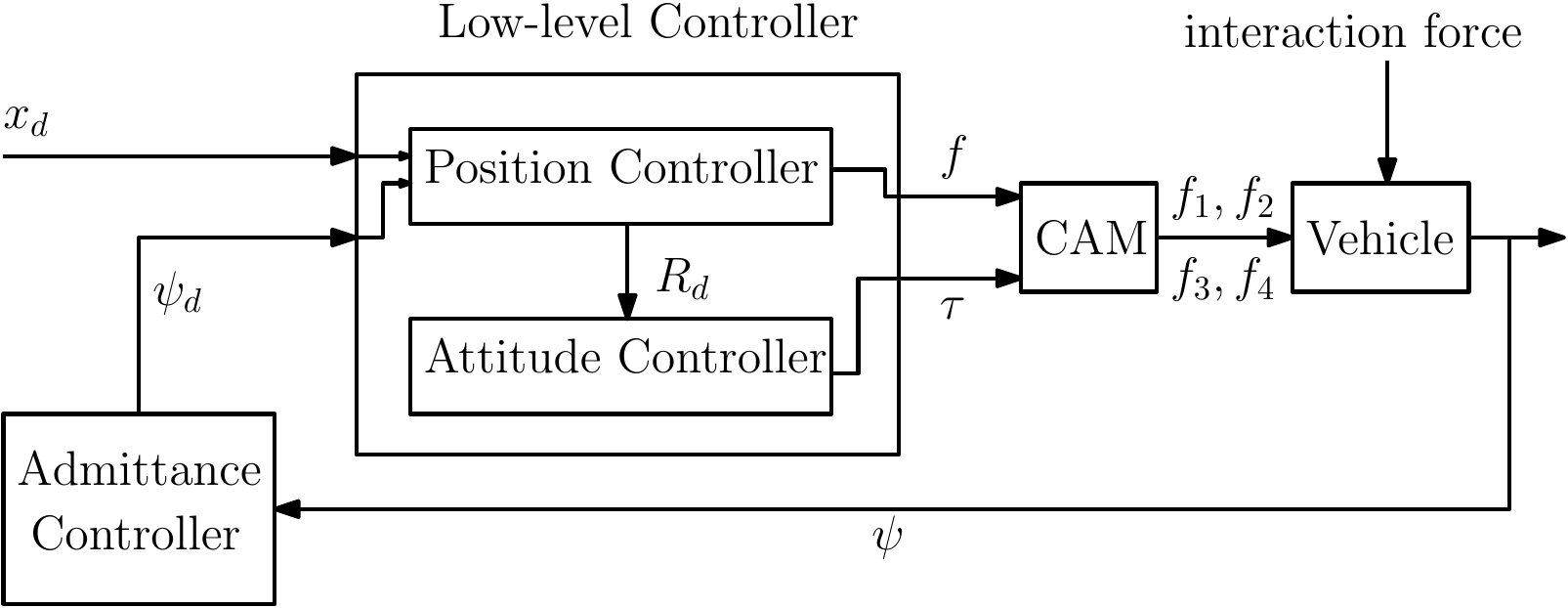}
    \caption{Block diagram of the SQUEEZE controller. The admittance control in yaw is outside the position and attitude control loops. The low-level controller consists of position and attitude control. In this work, we use a P-PID structure to compute input $f$ for the position loop and the proposed adaptive controller in (\ref{eqn:control_tau}) to generate input $\tau$.}
    \label{fig:cbd}
    \vspace{-0.1in}
\end{figure}

In this subsection, we propose an admittance controller in yaw to account for the physical forces acting on the quadrotor in relatively smaller gaps \revised{and} tunnels. It is critical to replan the yaw setpoint because as the quadrotor approaches the passageway, unforeseen interactions can lead to \revised{unintended} yaw moments.
In such scenarios, if the yaw set-point is not updated, the quadrotor tries to correct the yaw repeatedly during its flight and is \revised{prone} to multiple collisions \revised{which} may lead to unsuccessful flights. 
To this end, we propose a yaw admittance controller where an outer loop is added to the low level controller to modify the yaw reference trajectory. Figure \ref{fig:adm_schematic} demonstrates a scenario where a multicopter flies through a constrained space. Upon entering, the vehicle experiences interactions and the current yaw changes. At this instance, the vehicle would try to reject the interaction which could result a jerky motion. Therefore, an admittance controller is implemented to update the desired yaw according to the current yaw and comply with the interactions similar to the one in \cite{AA13}. The complete control structure is showed in Fig. \ref{fig:cbd}.
We use the control in (\ref{eqn:admittance}) to generate a new desired yaw
\begin{equation}\label{eqn:admittance}
    M_\psi \ddot{\psi_d} + D_\psi \dot{\psi_d}+ K_\psi \psi_d = \psi,
\end{equation}
where $\psi$ is the current yaw.
The tuning parameters of $M_\psi, D_\psi$ and $K_\psi$ are chosen such that the dynamics of $\psi_d$ is critically damped to track changes in current yaw, to avoid \revisedb{delayed responses and oscillations in $\psi_d$ from over-damped and under-damped dynamics, respectively.}

\section{Experiments and Discussions}\label{sec:experiments}
This section demonstrates the performance of the proposed system through experiments of i) flying through a gap and ii) flying through a passageway. 
\subsection{Hardware Setup}
Experiments were conducted in an indoor drone studio at the Arizona State University. An Optitrack motion capture system with 17 high-speed cameras \revisedb{was utilized to obtain the position and heading of the vehicle.} The 3-D position and current heading were transmitted to PIXHAWK \cite{M+11}, at 120 Hz for real-time feedback control. The high-level onboard computer was an Intel UP-board which ran the Robot Operating System (ROS) for communication with motion-capture system. A multi-threaded application was implemented for the admittance control algorithm and to enable the serial communication with the PIXHAWK. The multicopter system was equipped with two additional IMUs, one underneath each arm, to obtain the arm bending angle.

The low-level attitude control algorithm was implemented as described in Sec. \ref{sec:adaptivecontrol}. A quaternion-based complementary filter was implemented for attitude estimation. A Kalman filter based algorithm was implemented for the low-level position estimation, and a cascaded P-PID control structure for the position control module. The admittance parameters $(M_{\psi}, D_{\psi}, K_{\psi})$ were (0.01, 0.2, 1.0) for both the flight tests.

\begin{figure*}[t]
\centering
    \includegraphics[width=1\textwidth]{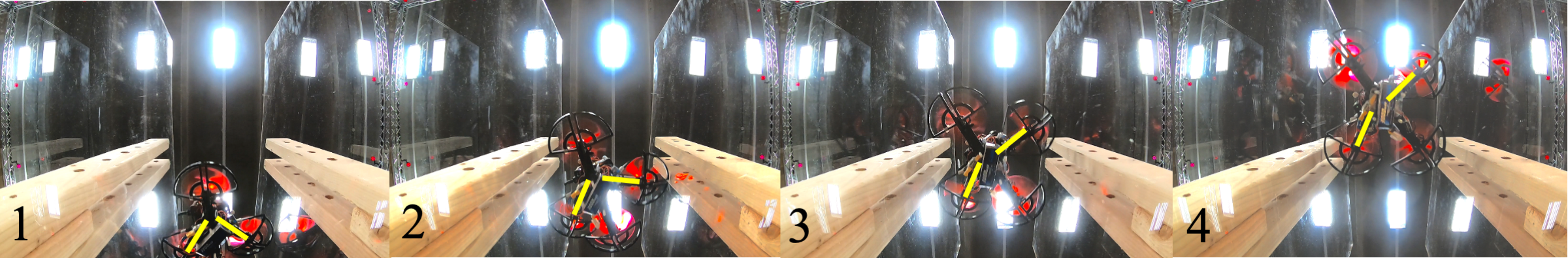}
    \caption{Bottom view of the SQUEEZE's flight via the passageway. The color marked arms show how physical interactions cause them to morph. (1) shows the quadcopter entering the passageway, (2-3) shows the quadcopter squeezing through the narrow passageway, and (4) shows the quadcopter exiting the passageway.}
    \label{fig:videosnap}
        \vspace{-0.1in}
\end{figure*}

\subsection{Flight through a Gap}
\begin{figure}[!b]
    \centering
    \subfloat[The SQUEEZE going through a gap]{
    \includegraphics[width=0.45\textwidth]{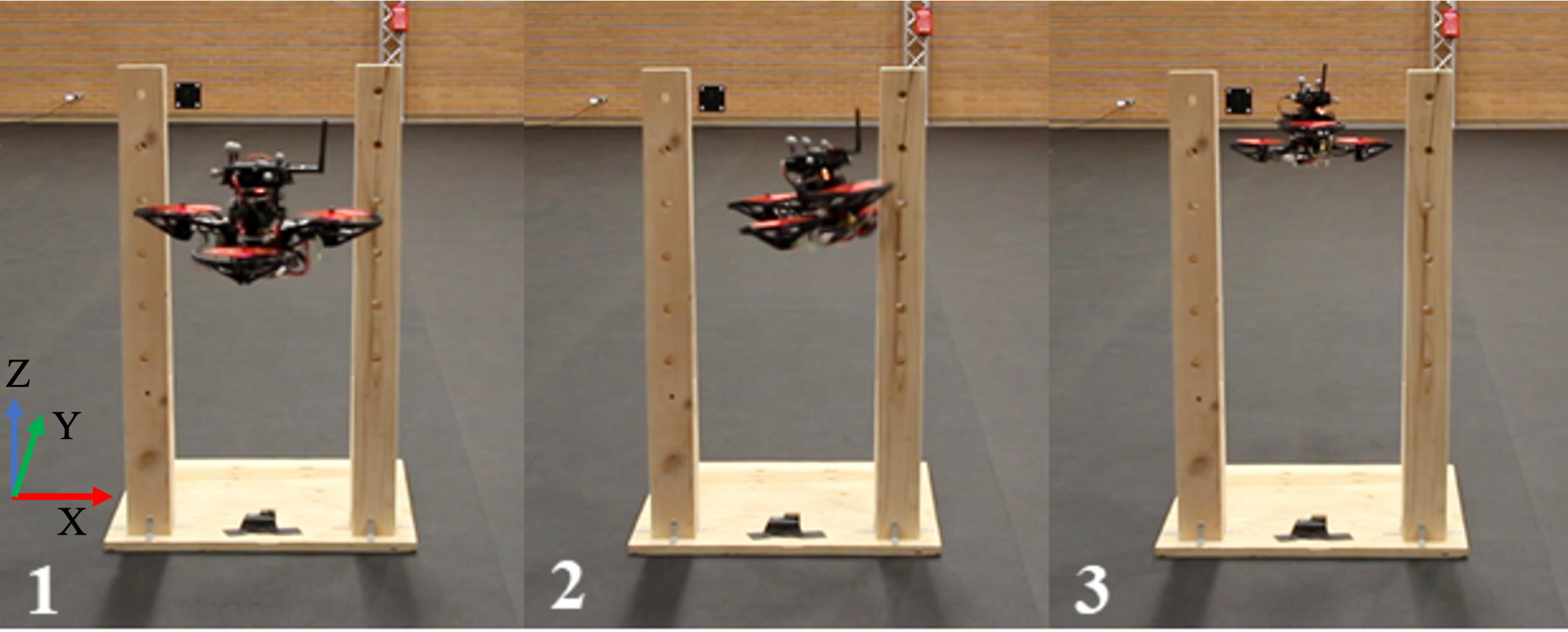}
    \label{fig:flightthroughgap}
  }\\
	\vspace{-0.1in}
  \subfloat[The SQUEEZE going through a passageway]{
    \includegraphics[width=0.45\textwidth]{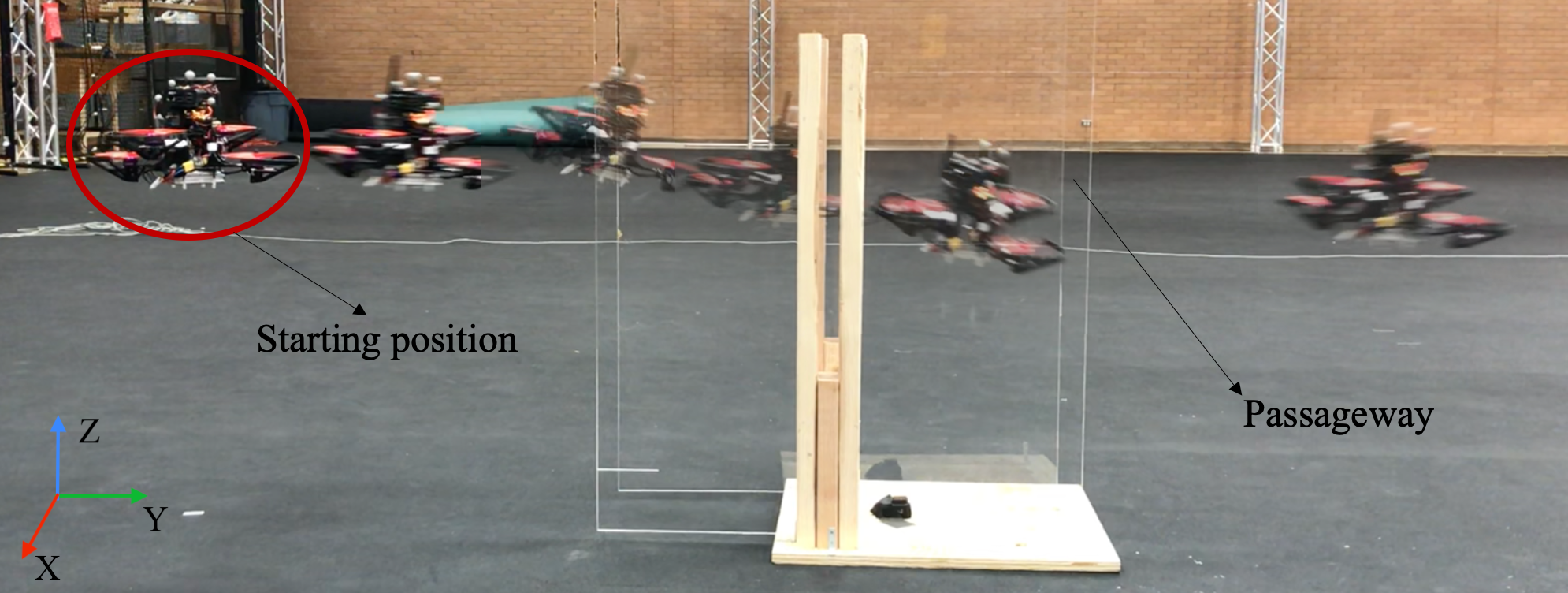} 
    \label{fig:flightthroughpass}}
    \caption{Snapshots of the SQUEEZE going through (a) a gap and (b) a passageway.}
    
\end{figure}

\begin{figure}
\centering
	\subfloat[{ 3D trajectory of the SQUEEZE with a 0.3$m$ height loss.  } ]
	{\includegraphics[width = 0.45\textwidth]{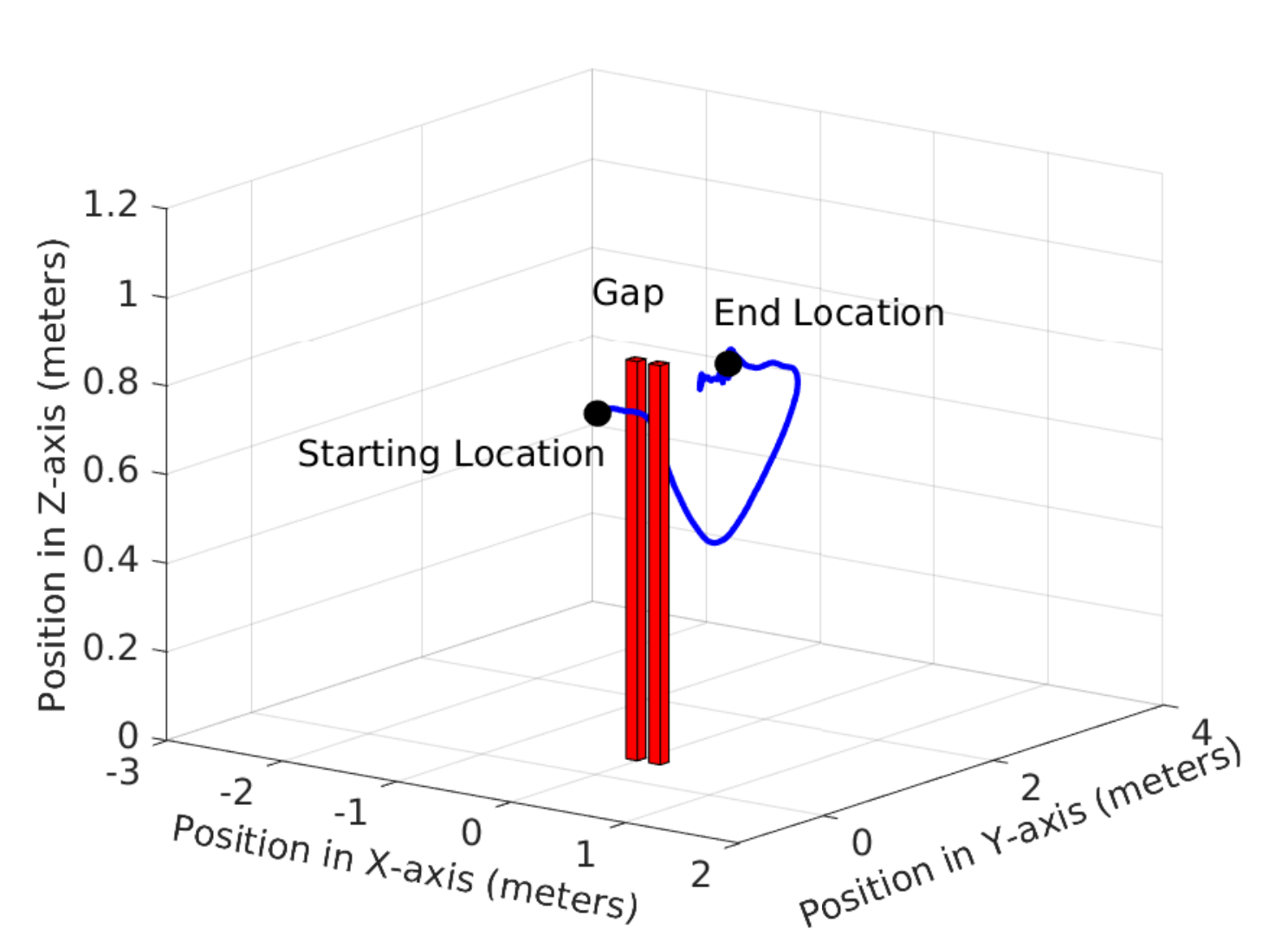}\label{fig:gap}
	}\\
	\subfloat[The desired yaw angle reference trajectory generated using (\ref{eqn:admittance}). ]
	{\includegraphics[width = 0.43\textwidth]{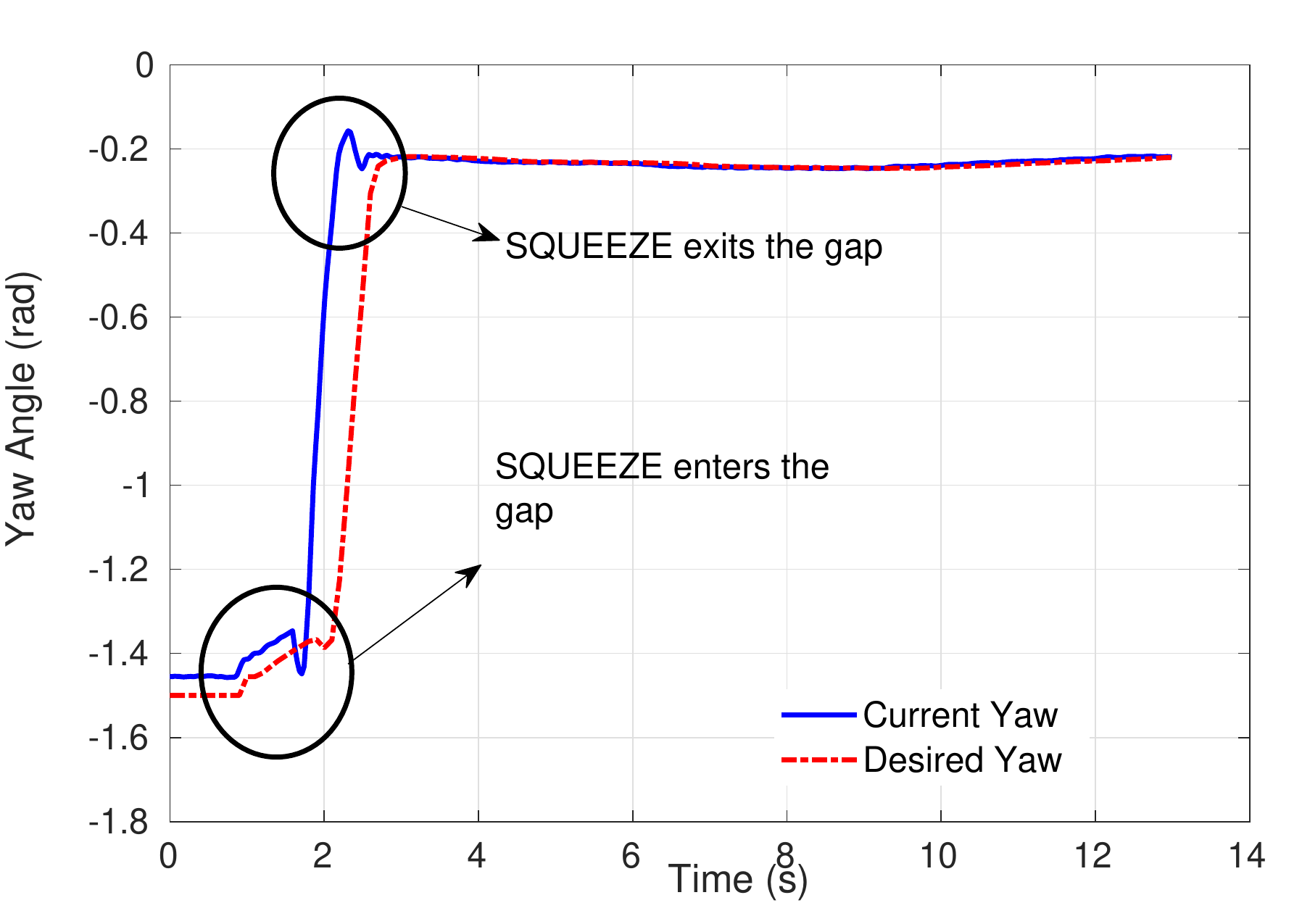}\label{fig:yawplotgap}
	}
	\caption{Results of a flight through a 0.29$m$$\times$0.04$m$ gap.  
}
    \label{fig:experimental_results}
    \vspace{-0.2in}
\end{figure}

In this flight test, the quadrotor autonomously flew through a gap. The gap was a pair of wooden beams, 90 cm tall, attached to a fixed base plate as shown in Fig. \ref{fig:flightthroughgap}. The two wooden beams were 30 cm apart, which was less than the width of the multicopter (41 cm) as measured from the tip of one propeller guard to the tip of the diametrically opposite propeller guard. The base of the gap was located at (0, 0.5, 0) and the tip's location was (0, 0.5, 0.9). The multicopter started at (0, 0, 0.8) and the target location was (0, 1.5, 0.8). Figure \ref{fig:gap} shows the trajectory of the SQUEEZE during flight. The interaction between the SQUEEZE and the gap lasts for about 0.8 seconds. It can be observed that, this interaction resulted in a loss of height by approximately 30 cm as well as a change in the heading (yaw) of the vehicle. The quadrotor complied to the change in the heading using the admittance controller as described in Section \ref{admittancectrl}. Figure \ref{fig:yawplotgap} demonstrates the current yaw and desired yaw of the system while in admittance mode. After the SQUEEZE entered the gap, the current yaw of the system changed due to interaction with the wooded beams. The admittance controller changed the yaw setpoint to comply with these interactions. After the SQUEEZE exited the gap, it returned to the height setpoint while maintaining the desired yaw generated by the admittance control. 


\subsection{Flight through a Passageway}
\begin{figure}
\centering
	\subfloat[{3D trajectory of the SQUEEZE with a 0.15$m$ height loss.} ]
	{\includegraphics[width = 0.45\textwidth]{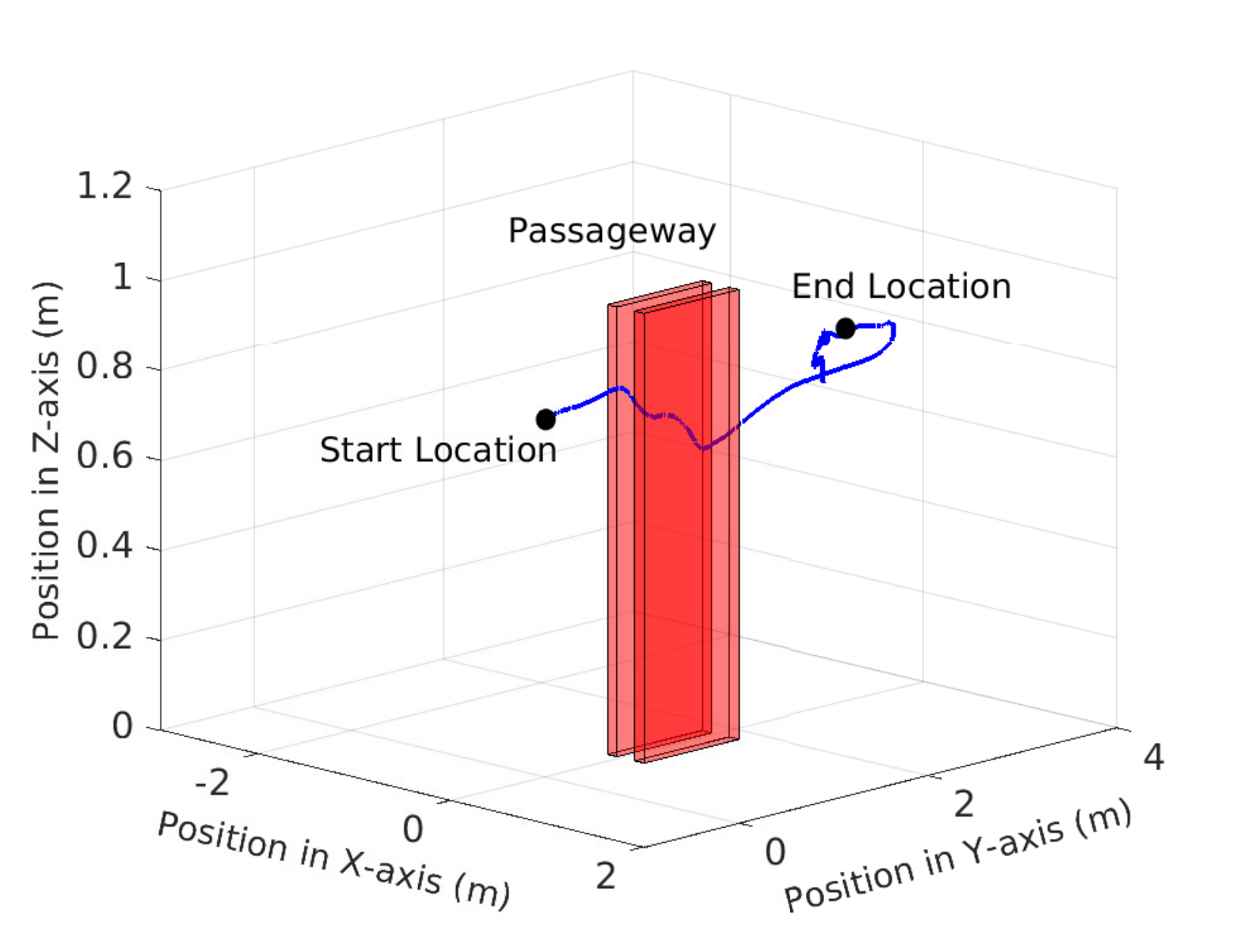}\label{fig:passageway}
	}\\
	\subfloat[{The desired yaw angle reference trajectory generated using (\ref{eqn:admittance}).} ]
	{\includegraphics[width = 0.45\textwidth]{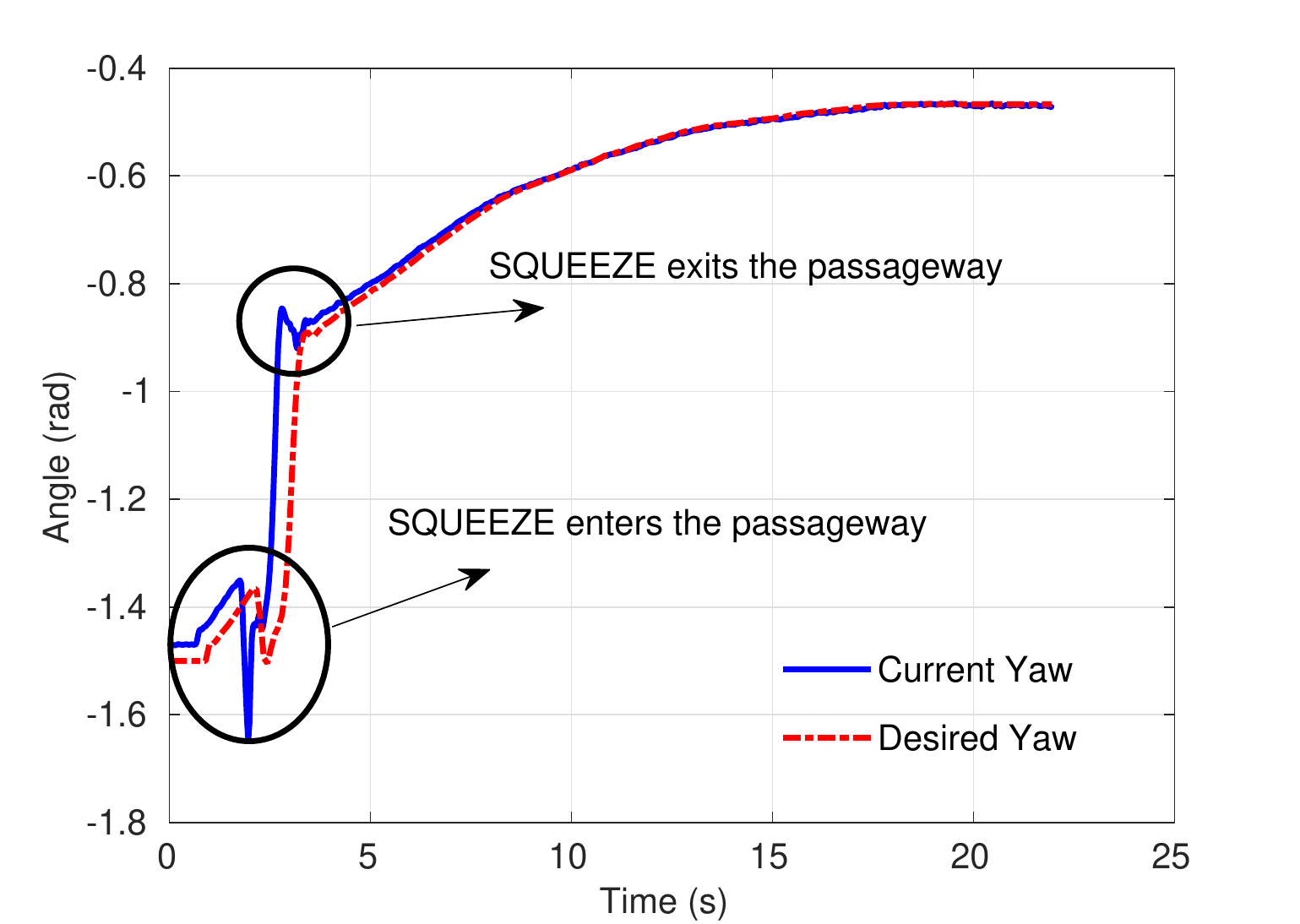}\label{fig:passagewayyaw}
	}
	\caption{Results of a flight through a 0.29$m$$\times$1$m$ passageway. }
    \label{fig:experimental_results2}
    \vspace{-0.2in}
\end{figure}

After successfully evaluating flight through a gap, the performance of the SQUEEZE is evaluated using a passageway where the vehicle continuously interacts with the environment. 
For the experimental setup, two acrylic sheets of 1 \textrm{m} length are utilized to create the passageway with a width of 29 \textrm{cm}. Figure \ref{fig:flightthroughpass} shows the video snapshots of the SQUEEZE flying through the passageway. The vehicle remains in the passageway for 1 second. The SQUEEZE started at (0, 0, 0.8) and the targeted location was (0, 3.0, 0.8). The beginning of the passageway is located at 80 cm from the origin along the $y$-axis. 
While successfully flying through the passageway, the SQUEEZE lost height by 15 cm. The loss in height can be attributed to potential near-wall aerodynamic effects\cite{walleffect} and physical interactions with the walls of the passageway. As shown in Figure \ref{fig:adm_schematic} and explained in Section \ref{admittancectrl}, an admittance control was implemented to change the desired yaw and comply with the interactions. Figure \ref{fig:passagewayyaw} represents the yaw admittance of the system under interactions. As the SQUEEZE enters the narrow passageway, the desired yaw changes according to the current yaw, which is the expected behaviour of admittance control. Figure \ref{fig:videosnap} depicts the arms morphing and adapting to the narrow passageway, while flying through it. 

\section{Conclusion and Future Work}\label{sec:RNC}
In this paper, we introduced the SQUEEZE, an novel quadrotor design with passive folding mechanism which could fly through gaps and passageways with dimensions smaller than its full body width while interacting with the environment. We developed an adaptive controller for trajectory tracking and 
employed a yaw admittance controller in the outer loop to account for physical interactions during the flights. The mechanical complexity and added weight of this design was low compared to existing morphing quadrotors. 
Finally, the proposed design was validated in flight tests through narrow apertures and tunnel-like environments. 

\revised{Future work includes designing new motion planning algorithms which can leverage its adaptive morphology. 
We also aim to study interaction with diffrent objects for improving robustness of the low-level control.} 

\begin{table*}[b!]
  \centering
  \small\addtolength{\tabcolsep}{-3pt}
  \caption{Validation of moment of inertia formulation (in  $kg-m^2$)}
  \label{table:moi}
    \begin{tabular}{ |c|>{\centering}p{3cm}|cccccc|cccccc| }
     \hline
     & \multirow{2}{*}{Configuration} & \multicolumn{6}{|c|}{Assumed Lumped Mass Model} & \multicolumn{6}{|c|}{SolidWorks} \\
      & & $J_{xx}$ & $J_{yy}$ & $J_{zz}$ & $J_{xy}$ & $J_{yz}$ & $J_{zx}$ & $J_{xx}$ & $J_{yy}$ & $J_{zz}$ & $J_{xy}$ & $J_{yz}$ & $J_{zx}$ \\ [1ex]
    \hline
     1 & $\beta_1 = 90^o$, $\beta_2 = 90^o$ & 0.0037 &0.0037 & 0.0067 & 0.0000 & 0.0000 & 0.0000 & 0.0036 & 0.0038 &0.0064 & 0.0000 & 0.0000 & 0.0000\\
     \hline
     2 & $\beta_1 = 45^o$, $\beta_2 = 45^o$ & 0.0023 &0.0050 & 0.0062 & 0.0000 & 0.0000 & -0.0002 & 0.0028 & 0.0042 &0.0060 & 0.0000 & 0.0000 & -0.0002 \\
     \hline
     3 & $\beta_1 = 60^o$, $\beta_2 = 30^o$ & 0.0022 &0.0050 & 0.0062 & -0.0001 & 0.0002 & 0.0001 & 0.0027 & 0.0042 &0.0059 & -0.0001 & 0.0002 & 0.0001 \\
     \hline
    \end{tabular}
\end{table*}
\bibliography{bibliography.bib}
\bibliographystyle{ieeetr}

\appendix

\textbf{Proof for attitude stability:}
We first find the error dynamics for $e_R , e_\Omega$, and define a Lyapunov function. Then, we show that with the proposed control, the attitude error is asymptotically stable to its zero equilibrium. The error function on $\mathsf{SO(3)}$ is chosen as \cite{L+10}:
\begin{equation}
    \Psi(R,R_d) = \frac{1}{2}\textrm{tr}[I - R_d^TR]
\end{equation}
From the definitions of $\Psi$ and $e_\Omega$, the derivatives of each quantity are
\begin{equation}\label{eqn:error_dynamics}
    \begin{aligned}
        \frac{d}{dt} (\Psi(R,Rd)) = &e_R \cdot e_\Omega \\
        \dot{e}_\Omega = &\dot{\Omega} - \zeta_d \\
        = &-k_R e_R - k_\Omega e_\Omega - k_{\Omega_d} \dot{e}_\Omega \\
        = &-k'_R e_R - k'_\Omega e_\Omega 
    \end{aligned}
\end{equation}
where $\dot{\Omega}$ is the angular acceleration of the quadrotor measured in the body frame, 
and $k'_R, k'_\Omega, k'_{\Omega_i}$ are positive constants. 
\par 
Now, consider the following Lyapunov candidate
\begin{equation}
    \mathcal{V} = \frac{1}{2} e_\Omega \cdot e_\Omega + k'_R \Psi(R,Rd)
\end{equation}

Using (\ref{eqn:dynamics}), (\ref{eqn:control_tau}) and (\ref{eqn:error_dynamics}), the time derivative of $\mathcal{V}$ follows:
\begin{equation}
    \begin{aligned}
        \dot{\mathcal{V}}~ = ~& e_\Omega \cdot \dot{e}_\Omega + k_R e_R \cdot e_\Omega \\
        ~=~ & e_\Omega (-k'_R e_R - k'_\Omega e_\Omega) + k'_R e_R \cdot e_\Omega \\
        ~=~ & -k'_\Omega \norm{e_\Omega}^2 
    \end{aligned}
\end{equation}
Since $k'_\Omega$ is a positive constant, we see that $\dot{\mathcal{V}}$ is negative, therefore proving asymptotic stability of the zero equilibrium for the attitude tracking. 
This completes the proof.  $\square$

\end{document}